\pdfoutput=1
\documentclass{article}
\usepackage[final]{colm2026_conference}

\usepackage{microtype}
\usepackage{hyperref}
\usepackage{url}
\usepackage{booktabs}
\usepackage{graphicx}
\usepackage{amsmath}
\usepackage{array}
\usepackage{xcolor}
\usepackage{caption}

\definecolor{darkblue}{rgb}{0, 0, 0.5}
\hypersetup{colorlinks=true, citecolor=darkblue, linkcolor=darkblue, urlcolor=darkblue}

\graphicspath{{figures/}}
\title{Targeting the Attention Heads Behind \\ Object Hallucination in LLaVA}

\author{Armaan Sandhu\textsuperscript{*}, Abhilasha Senapati\textsuperscript{*} \& Hima Kammachi\textsuperscript{*} \\
University of Massachusetts Amherst \\
\texttt{\{apsandhu, asenapati, hkammachi\}@umass.edu} \\
\textsuperscript{*}Equal contribution.}

\begin{document}

\maketitle
\lhead{\footnotesize Accepted at the Actionable Interpretability Workshop, COLM 2026}

\begin{abstract}
Vision-language models such as LLaVA-1.5-7B often hallucinate objects absent
from the image when generating captions. We ask whether an interpretability
diagnosis of this failure can guide a targeted fix, and we measure what that
fix actually changes. We rank attention heads by how much their image attention
drops around hallucinated object words, then  screen the shortlist by
ablating candidate heads and measuring the change in hallucination-token log
probability, yielding a 32-head set. We restrict two interventions to these
heads: a head-sliced LoRA adapter and an inference-time grounding controller.
On 400 held-out COCO images, the combined method lowers CHAIRs (the fraction of
captions with a hallucinated object) from $0.370$ to $0.230$ and CHAIRi (the
fraction of hallucinated object mentions) from $0.156$ to $0.096$ ($p < 0.001$,
paired sign-flip tests). Two controls sharpen attribution. A random-head LoRA
control, matched layer-for-layer and trained identically, performs no better
than the matched baseline on a separate 200-image control split, supporting the
role of head selection rather than LoRA capacity. Under fixed decoding budgets,
the CHAIR reduction persists and grows with budget ($23\%$ at 64 tokens to
$58\%$ at 128), arguing against a pure max-token or truncation artifact, although
the method remains shorter and more conservative. The resulting behavior reduces
unsupported object mentions while also lowering object recall ($0.78$ to $0.70$). We present a diagnosis-to-intervention pipeline for object hallucination, and, more importantly, a controlled account of what acting on the diagnostic signal actually does: it localizes intervention sites with real, non-random leverage, reported as a behavioral profile rather than a single score.
\end{abstract}

\section{Introduction}
Vision-language models often produce fluent captions that mention objects not
present in the image. This object hallucination failure undermines reliability
in any setting where generated descriptions feed downstream decisions
\citep{rohrbach2018chair}. It also makes a useful test case for
interpretability: the behavior can be localized to specific model components,
measured against ground-truth annotations, and then used to ask a sharper
question than whether an internal signal merely correlates with the failure,
namely whether acting on that signal actually changes behavior.

Existing mitigation methods usually intervene either at decoding time or through
broad model adaptation. Training-free decoding methods such as
VCD~\citep{leng2024vcd} and SPIN~\citep{sarkar2025spin} adjust the output
distribution or suppress low-image-attention heads at inference time.
Parameter-efficient fine-tuning can shift model behavior with small updates~\citep{hu2022lora,dettmers2024qlora}, but
is rarely restricted to components implicated by an object-level diagnosis. We
instead use an interpretability diagnosis to decide both \emph{where} to adapt
and \emph{when} to intervene: the same ablation-screened head set determines a
head-restricted LoRA adapter and an inference-time grounding controller.

We study LLaVA-1.5-7B~\citep{liu2023llava, liu2024improved}. Our pipeline ranks attention heads
by how much their attention to the image drops around hallucinated object words, screens the strongest candidates by ablating them and measuring the
change in hallucination-token log probability, and arrives at a 32-head set. We
then apply two interventions limited to those heads: a LoRA adapter on the
selected heads, and a decoding-time grounding penalty that fires when those
heads attend weakly to image tokens.

The combined intervention is conservative: it produces shorter captions with
lower object recall. Its CHAIR gains therefore reflect two things at once,
genuine hallucination reduction and a general shift toward saying less. We
measure both and report the intervention as a behavioral profile rather than a
single score. 

Our contributions are:
\begin{itemize}
\item \textbf{We introduce a diagnosis-to-intervention pipeline for object
hallucination.} The pipeline identifies hallucination-linked attention heads,
screens them by ablation, and restricts both a LoRA adapter and an
inference-time grounding controller to the resulting 32-head set.

\item \textbf{We show that head selection matters beyond LoRA capacity.} On a
separate 200-image control split, a layer-matched random-head LoRA control
trained with the identical recipe performs no better than the matched baseline
(CHAIRs $0.400$), while the selected-head adapter reduces CHAIRs to $0.100$
on the same images.

\item \textbf{We reduce object hallucination on held-out COCO images.}
On 400 held-out images, the full LoRA + Grounding method reduces CHAIRs from
$0.370$ to $0.230$ and CHAIRi from $0.156$ to $0.096$, with paired sign-flip
tests showing significant reductions ($p < 0.001$).

\item \textbf{We test robustness to decoding length.} With the same token
budgets, our method still reduces CHAIRs, with relative gains growing
from $23\%$ at 64 tokens to $58\%$ at 128 tokens. This rules out a simple
max-token truncation explanation, though the method remains shorter and more
conservative.

\item \textbf{We characterize the intervention as a behavioral profile rather
than a single score.} The method reduces unsupported object mentions but also
shortens captions and lowers object recall, so we report hallucination,
length, recall, random-head controls, SPIN/VCD comparisons, and sensitivity
checks together.
\end{itemize}

\section{Related Work}
LLaVA-style models combine visual encoders with instruction-tuned language
models and achieve strong open-ended multimodal generation~\citep{liu2023llava, liu2024improved}.
However, they can inherit language-prior behavior that produces visually
unsupported details. CHAIR formalizes object hallucination in image captioning
by measuring whether generated object mentions appear in the image
annotations~\citep{rohrbach2018chair}. Although CHAIR is an incomplete proxy for
visual truth, it gives a concrete object-level failure signal suitable for
controlled intervention studies.

\paragraph{Attention attribution for hallucination.}
A growing body of interpretability work links object hallucination to where
LVLMs place attention. \citet{jiang2025devils} use an attention lens to localize
hallucination to the middle layers of LLaVA-style models and show that grounded
object tokens receive higher visual attention than hallucinated ones.
\citet{liu2024pai} find that image tokens receive little attention during
generation and intervene on decoder self-attention to make inference more
image-centric. Most directly related, \citet{yang2025modular} use causal
mediation analysis to identify ``hallucination heads'' in the middle and deeper
layers that over-attend to text, and mitigate hallucination with both a
decoding-time adjustment and a head-targeted fine-tuning method.

Our pipeline shares this diagnose-then-target strategy and the use of both a
decoding and a training intervention. We differ in what we measure rather than
in the targeting idea: we add a layer-matched random-head control that isolates
the contribution of head \emph{selection} from adapter capacity, a fixed-budget
analysis that separates hallucination reduction from max-token truncation
effects, and an explicit account of the precision-coverage tradeoff the
intervention induces. Where prior work reports mitigation, we report what acting
on the diagnostic signal does as a full behavioral profile.

\paragraph{Targeted adaptation and training-free decoding.}
Parameter-efficient adaptation methods such as LoRA and QLoRA show that model
behavior can often be shifted with small low-rank updates rather than full
fine-tuning~\citep{hu2022lora,dettmers2024qlora}. Preference objectives such as
DPO provide a simple way to train from chosen and rejected
responses~\citep{rafailov2023dpo}. Our work combines these ideas with a more
localized target: only heads implicated by the hallucination diagnosis are
allowed to change.

Training-free decoding methods provide complementary comparisons for the
inference-time part of our pipeline. Visual
Contrastive Decoding (VCD) contrasts output distributions under real and
noise-corrupted images to suppress language-prior-driven
tokens~\citep{leng2024vcd}. SPIN suppresses dynamically selected
low-image-attention heads at inference time~\citep{sarkar2025spin}, and is our
closest decoding baseline since it also uses attention to choose heads. We
differ from SPIN in two ways: we use a fixed, ablation-screened head set rather
than re-selecting heads at each step, and we train a head-restricted adapter on
that set rather than only suppressing at inference. VCD instead applies a
distributional contrast between real and corrupted images, without explicitly
selecting hallucination-prone heads.

\section{Method}
\subsection{Task and Model}
We evaluate open-ended image captioning with \texttt{llava-hf/llava-1.5-7b-hf}.
All main experiments use COCO val2014 images and object
annotations~\citep{lin2014coco}. The prompt is held fixed across methods:
``Describe this image in detail.'' We report CHAIRs, the fraction of captions
with at least one hallucinated object, and CHAIRi, the fraction of object
mentions that are hallucinated.

Our pipeline has three sequential stages: head selection by diagnosis
(Section~\ref{sec:diagnosis}), a LoRA adapter trained on the selected heads, and
a grounding controller applied on top of the adapter over the same head set. We
evaluate the LoRA stage and the full \emph{LoRA + Grounding} method against the
greedy baseline, and we also report a \emph{Grounding only} condition (the
controller applied to the base model, without LoRA) as a reference that isolates
the controller's effect; it is not part of the pipeline.

\subsection{Object-Level Diagnosis}
\label{sec:diagnosis}
We generate captions on a screening set of 200 COCO val2014 images and label
each object mention as grounded or hallucinated with a COCO-derived CHAIR
scorer. For every object token we record per-head attention to image tokens
across the 1024 language-model attention heads of LLaVA-1.5-7B (32 layers, 32
heads). Heads are ranked by how much their image attention decreases around
hallucinated object words relative to grounded object words, producing a fast
diagnostic shortlist.

We then ablation-screen the strongest candidates: we ablate heads one at a time
and measure the change in hallucination-token log probability. To avoid
selecting on the same signal twice, the screen runs over both shortlisted and
non-shortlisted heads, so a candidate can be rejected if its ablation effect is
small. We retain the 32 heads with the largest ablation effect, spanning 19
transformer layers. This set is fixed and used by both later stages.

\subsection{LoRA Adaptation}
The LoRA stage updates only the layers and head slices associated with the 32
selected heads. LoRA modules are attached to the Q/K/V projections in layers
containing selected heads, and gradient masking restricts learning to the
selected head dimensions. Training examples are contrastive caption pairs: the
chosen response is a COCO ground-truth caption and the rejected response is the
model's own hallucinated caption. We optimize a DPO-style objective~\citep{rafailov2023dpo} with rank
$r=8$, $\alpha=16$, dropout $0.05$, for three epochs. Because the chosen
responses are COCO captions, which are shorter than the model's free-running
output, this objective also biases the adapter toward brevity; we analyze the
resulting length effect in Section~\ref{sec:budget-robustness}.

\subsection{Inference-Time Grounding}
The grounding controller complements the LoRA adapter rather than operating as a
standalone intervention. It applies a conservative guard at inference time using
the same head set that guides adaptation.

At each generation step it computes a grounding score from the visual attention
mass assigned by the selected heads:
\[
g_t =
\mathrm{mean}_{h \in H}
\frac{\sum_{i \in \mathrm{img}} \alpha_{h,i}}
{\sum_j \alpha_{h,j}}.
\]
Here $H$ is the ablation-screened 32-head set, $i$ indexes image token
positions, $j$ indexes all attended positions at the current step (image,
prompt, and previously generated tokens), and $\alpha_{h,i}$ is the attention
weight from the current step to position $i$ under head $h$. Thus $g_t$ is the
fraction of head-$H$ attention mass placed on image tokens.

When $g_t$ falls below a threshold $\theta$, object-word logits from a
COCO-derived object vocabulary receive a penalty:
\[
\ell_v \leftarrow \ell_v + \lambda\,\min(0,\, g_t-\theta)
\quad \text{for object token } v,
\]
with $\lambda > 0$, so the penalty activates only when head-$H$ image attention
is below threshold and is inactive otherwise. Main results use $\theta=0.08$ and $\lambda=10$. The controller is deliberately lightweight: it does not prove
grounding at each step, but operationalizes the diagnostic signal as a
conservative guard against object mentions when the selected heads attend weakly
to the image. The full method applies this controller on top of the LoRA
adapter, over the same head set.

\begin{figure}[t]
  \centering
  \includegraphics[width=\linewidth]{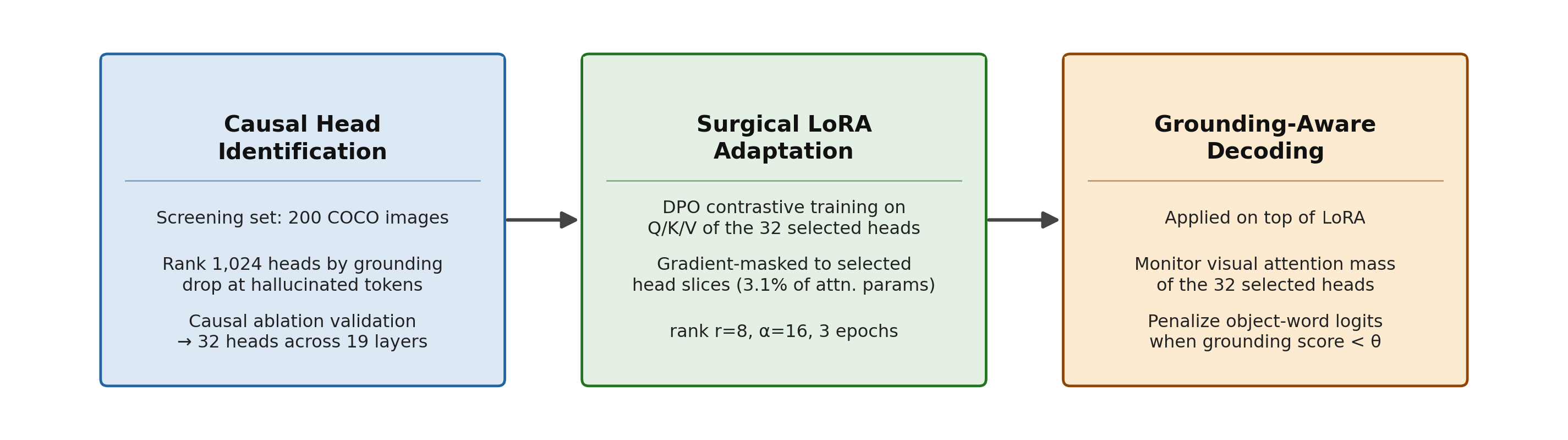}
  \caption{Diagnosis-to-intervention pipeline. Head selection identifies
  hallucination-linked heads, the LoRA stage adapts only those head slices, and
  the grounding controller is applied on top over the same head set.}
  \label{fig:pipeline}
\end{figure}

\section{Experiments}
\subsection{Protocol}
The main evaluation uses 400 held-out COCO val2014 images and four conditions:
the greedy LLaVA baseline, LoRA, Grounding only, and the full LoRA + Grounding
method. We report bootstrap 95\% confidence intervals over images and paired
sign-flip permutation tests against the baseline on the same images.

We also compare all conditions against SPIN and VCD under fixed decoding budgets
of 64, 80, and 128 new tokens on the same 400 images. Regenerating every method
under the same budget controls for the strong effect of caption length on CHAIR.
VCD uses its published hyperparameters ($\alpha=1.0$, $\beta=0.1$, noise step
500), implemented as a custom autoregressive loop on the same HF model for
identical tokenization and prompt handling~\citep{leng2024vcd}. All fixed-budget
results appear in one table (Table~\ref{tab:budget_all}).

Finally, we run a head-selection control on a 200-image held-out split. It uses
the same DPO pairs, LoRA configuration, and training recipe as our LoRA stage,
but replaces the ablation-screened 32 heads with a random set matched to the
same per-layer head counts, excluding the selected heads. We compare CHAIR on
the same images.

\subsection{Main Results}
Table~\ref{tab:main_bootstrap} shows that LoRA alone reduces hallucination
significantly, and the grounding controller reduces it further when
applied on top; the full method performs best. Grounding alone does not produce
a significant CHAIRi reduction (Table~\ref{tab:main_paired_deltas}, $p=0.2096$),
consistent with its design as a modifier rather than a standalone intervention.
The full method reduces CHAIRs by 37.8\% and CHAIRi by 38.5\% relative to the
baseline. This headline comparison is not length-matched, since the method also
shortens captions substantially; we therefore read it as a conservative
intervention that changes the targeted failure mode, and defer broader
caption-quality claims to length-matched and human evaluations.

We next present the head-selection control, the most direct test of whether the
diagnostic signal, rather than LoRA capacity, drives the change.

\begin{table}[t]
\centering
\caption{Main 400-image held-out comparison with bootstrap 95\% confidence intervals. Lower CHAIRs and CHAIRi are better.}
\label{tab:main_bootstrap}
\resizebox{\textwidth}{!}{%
\begin{tabular}{lccc}
\toprule
Method & CHAIRs & CHAIRi & Avg. len. \\
\midrule
Baseline & 0.3700 [0.3225, 0.4175] & 0.1558 [0.1327, 0.1800] & 61.1 \\
LoRA & 0.2650 [0.2225, 0.3075] & 0.1043 [0.0857, 0.1245] & 27.6 \\
Grounding only & 0.3100 [0.2650, 0.3575] & 0.1407 [0.1174, 0.1654] & 62.1 \\
LoRA + Grounding (Ours) & \textbf{0.2300 [0.1900, 0.2700]} & \textbf{0.0958 [0.0761, 0.1163]} & 29.6 \\
\bottomrule
\end{tabular}%
}
\end{table}

\begin{table}[t]
\centering
\caption{Paired deltas versus the baseline on the same 400 held-out images. Negative deltas are better. P-values use paired sign-flip permutation tests with 10000 resamples.}
\label{tab:main_paired_deltas}
\resizebox{\textwidth}{!}{%
\begin{tabular}{lcc}
\toprule
Method & $\Delta$ CHAIRs & $\Delta$ CHAIRi \\
\midrule
LoRA & -0.1050 [-0.1575, -0.0550], $p=0.0002$ & -0.0515 [-0.0738, -0.0295], $p=0.0001$ \\
Grounding only & -0.0600 [-0.1125, -0.0075], $p=0.0287$ & -0.0151 [-0.0386, +0.0082], $p=0.2096$ \\
LoRA + Grounding (Ours) & \textbf{-0.1400 [-0.1900, -0.0900], $p=0.0001$} & \textbf{-0.0601 [-0.0837, -0.0369], $p=0.0001$} \\
\bottomrule
\end{tabular}%
}
\end{table}

\subsection{Random-Head Control}
\label{sec:random-head}
On the 200-image control split, the layer-matched random-head LoRA control is
indistinguishable from the baseline (CHAIRs 0.400, CHAIRi 0.131), while the
selected-head adapter reduces CHAIRs to 0.100 and CHAIRi to 0.054 on the same
images. Replacing the selected heads with a layer-matched random set does not
reproduce the adapter's behavioral change, so the diagnostic signal identifies
intervention sites with distinct leverage. This control does not by itself
establish improved grounding, since the adapter's reduction remains subject to
the length and conservatism effects we examine next.

\subsection{Tradeoff: Fewer Hallucinations, Shorter Captions}
\label{sec:tradeoff}
\begin{figure}[t]
  \centering
  \includegraphics[width=\linewidth]{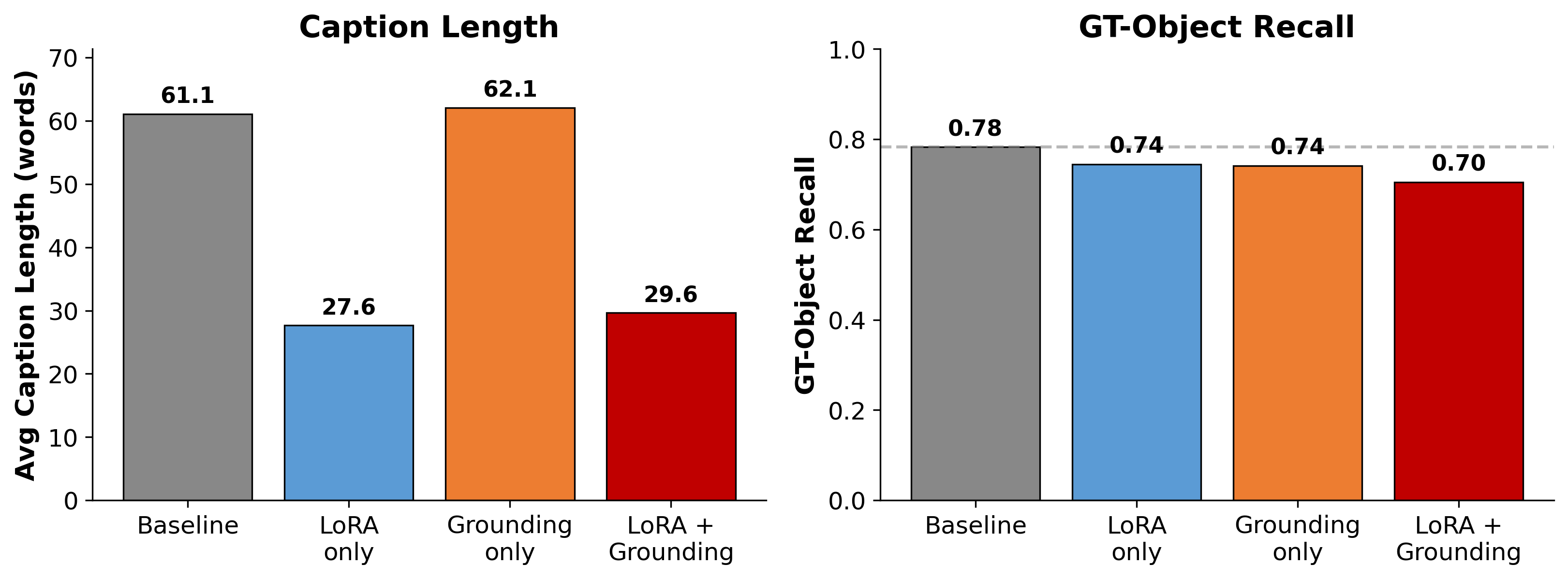}
  \caption{Hallucination reduction is accompanied by shorter captions and lower object recall. This tradeoff is central to evaluating actionability.}
  \label{fig:length-recall}
\end{figure}

The full method is substantially more conservative than the baseline: average
caption length drops from 61.1 to 29.6 words and object recall from 0.78 to 0.70
(Figure~\ref{fig:length-recall}). This matters because CHAIR can be reduced by
mentioning fewer objects. Our reading is therefore conditional: the method
reduces object hallucination under CHAIR but does not uniformly improve caption
quality. The fixed-budget analysis below addresses the strongest form of this
concern; a fully length-matched evaluation with human quality assessment remains
the most important missing control.

\subsection{Fixed-Budget Comparison and Robustness}
\label{sec:budget-robustness}
Table~\ref{tab:budget_all} reports all methods at fixed decoding budgets of 64,
80, and 128 new tokens on the same 400 images, with bootstrap 95\% confidence
intervals for Baseline, SPIN, and VCD. It serves two purposes: comparison
against training-free baselines under a controlled length budget, and a test of
whether our CHAIR reduction is a shorter-caption artifact.

\textbf{Comparison to SPIN and VCD.} Neither training-free baseline consistently
reduces hallucination: SPIN is numerically worse than its matched baseline at 64
and 80 tokens, and VCD increases CHAIRi at all three budgets. At 128 tokens SPIN
improves CHAIRs but leaves CHAIRi essentially unchanged. We read these as
protocol-specific comparisons, not general claims about SPIN or VCD.

\begin{figure}[t]
  \centering
  \includegraphics[width=\linewidth]{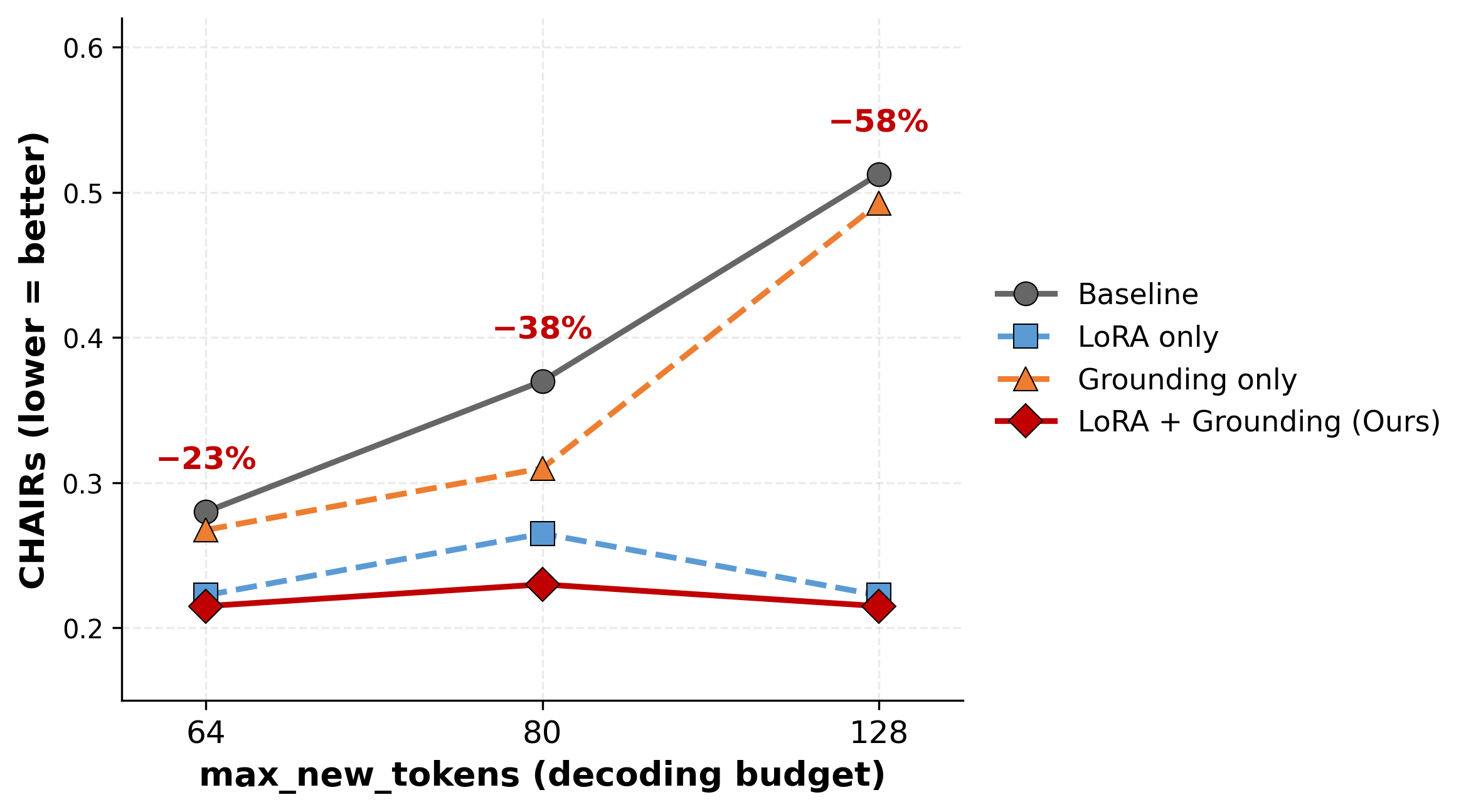}
  \caption{Caption-level CHAIR versus decoding budget on 400 held-out COCO images. Baseline CHAIRs rises sharply (0.28 $\to$ 0.37 $\to$ 0.51) as the budget grows, while LoRA + Grounding (Ours) stays near 0.22 at all budgets. The relative reduction grows from 23\% at 64 tokens to 58\% at 128, indicating the baseline becomes progressively miscalibrated to the image under larger budgets while the targeted intervention does not.}
  \label{fig:budget-calibration}
\end{figure}

\textbf{Robustness to caption length.} Within a fixed budget, both methods share
the same maximum length, removing the free-running length confound. At a
64-token budget that forces baseline truncation (baseline length 61.1 to 49.5
words), our method still reduces CHAIRs from 0.280 to 0.215, a 23.2\% relative
reduction. The gain therefore is not purely a max-token or truncation artifact, though the
method still generates shorter captions within the same budget. The
gap also \emph{grows} with budget: baseline CHAIRs rises from 0.280 to 0.512 as
the budget triples, while our method stays near 0.215 throughout, so the relative
reduction grows from 23\% at 64 tokens to 58\% at 128
(Figure~\ref{fig:budget-calibration}). We read this as the baseline becoming
progressively miscalibrated to the image as more tokens must be filled, a drift
the targeted intervention does not show. This does not achieve full length
matching, since our method still generates shorter captions within each budget
(27.9 vs.\ 49.5 words at 64 tokens); what it rules out is the strong claim that
the entire reduction comes from generating less.

\begin{table}[t]
\centering
\caption{All methods at fixed decoding budgets on the same 400 held-out images. Baseline, SPIN, and VCD include bootstrap 95\% confidence intervals; our conditions report point estimates, with bootstrap intervals for these fixed-budget runs left to appendix evaluation. Lower CHAIRs and CHAIRi are better. LoRA + Grounding (Ours) outperforms both training-free baselines at every budget; the relative reduction over the matched baseline grows from 23\% at 64 tokens to 58\% at 128 tokens.}
\label{tab:budget_all}
\resizebox{\textwidth}{!}{%
\begin{tabular}{c l c c c}
\toprule
Budget & Method & CHAIRs & CHAIRi & Avg.\ len. \\
\midrule
64 & Baseline                & 0.2800 [0.2375, 0.3225] & 0.1066 [0.0863, 0.1266] & 49.5 \\
64 & SPIN                    & 0.3025 [0.2575, 0.3475] & 0.1216 [0.1000, 0.1445] & 48.8 \\
64 & VCD                     & 0.2900 [0.2475, 0.3350] & 0.1301 [0.1069, 0.1559] & 48.8 \\
64 & LoRA                    & 0.2225                  & 0.0940                  & 28.4 \\
64 & Grounding only          & 0.2675                  & 0.1053                  & 49.3 \\
64 & LoRA + Grounding (Ours) & \textbf{0.2150}         & \textbf{0.0909}         & 27.9 \\
\midrule
80 & Baseline                & 0.3650 [0.3175, 0.4125] & 0.1382 [0.1155, 0.1601] & 60.8 \\
80 & SPIN                    & 0.3675 [0.3200, 0.4150] & 0.1419 [0.1194, 0.1657] & 58.4 \\
80 & VCD                     & 0.3850 [0.3375, 0.4325] & 0.1479 [0.1255, 0.1719] & 60.4 \\
80 & LoRA                    & 0.2650                  & 0.1043                  & 27.6 \\
80 & Grounding only          & 0.3100                  & 0.1407                  & 62.1 \\
80 & LoRA + Grounding (Ours) & \textbf{0.2300}         & \textbf{0.0958}         & 29.6 \\
\midrule
128 & Baseline                & 0.5125 [0.4650, 0.5600] & 0.1794 [0.1565, 0.2040] & 84.2 \\
128 & SPIN                    & 0.4750 [0.4250, 0.5225] & 0.1786 [0.1551, 0.2033] & 78.2 \\
128 & VCD                     & 0.5100 [0.4625, 0.5575] & 0.1951 [0.1709, 0.2214] & 84.0 \\
128 & LoRA                    & 0.2225                  & 0.0947                  & 45.0 \\
128 & Grounding only          & 0.4925                  & 0.1734                  & 83.9 \\
128 & LoRA + Grounding (Ours) & \textbf{0.2150}         & \textbf{0.0917}         & 43.7 \\
\bottomrule
\end{tabular}%
}
\end{table}

\subsection{Sensitivity Checks}
\begin{figure}[t]
  \centering
  \includegraphics[width=\linewidth]{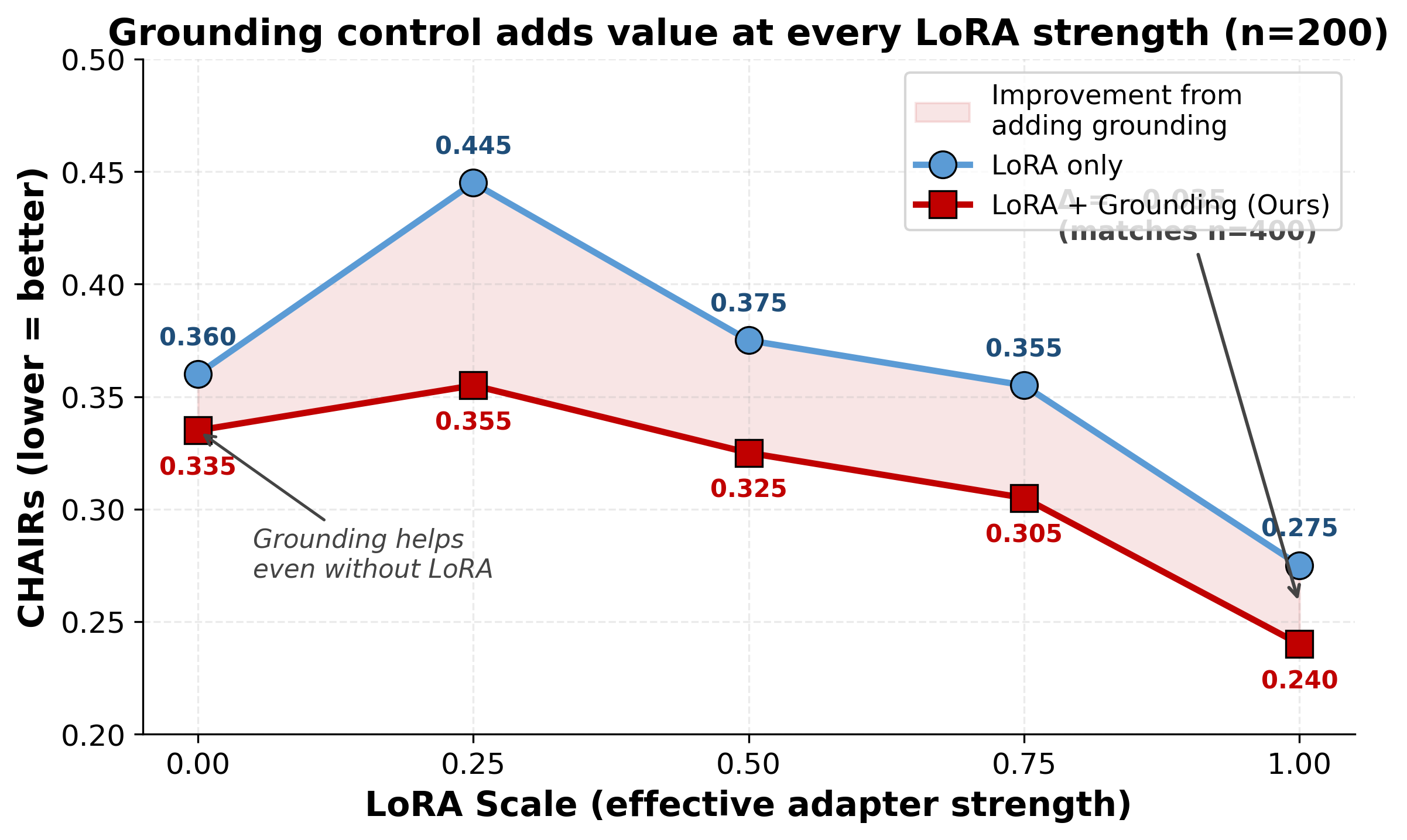}
  \caption{Adapter-scale sweep on 200 held-out images. Grounding control improves CHAIRs across LoRA strengths, but does not resolve the separate length-matching and head-selection questions.}
  \label{fig:lora-scale}
\end{figure}

Figure~\ref{fig:lora-scale} shows an adapter-scale sweep on 200 held-out images:
the grounding controller improves CHAIRs at every LoRA scale, including the
no-adapter and full-adapter settings, so it contributes beyond the LoRA update
even though its standalone 400-image CHAIRi reduction is not significant. The
controller is also insensitive to threshold over $\theta \in
\{0.04,0.06,0.08,0.10,0.12\}$ (CHAIRs stays in 0.24--0.26), and top-16 versus
top-32 heads gives nearly identical CHAIRs (0.24) and CHAIRi (0.099 vs.\ 0.098).
With the random-head control (Section~\ref{sec:random-head}), these checks
support the head-selection story but do not substitute for length-matched
400-image controls.

\subsection{Qualitative Example}

\begin{figure}[t]
  \centering
  \includegraphics[width=\linewidth]{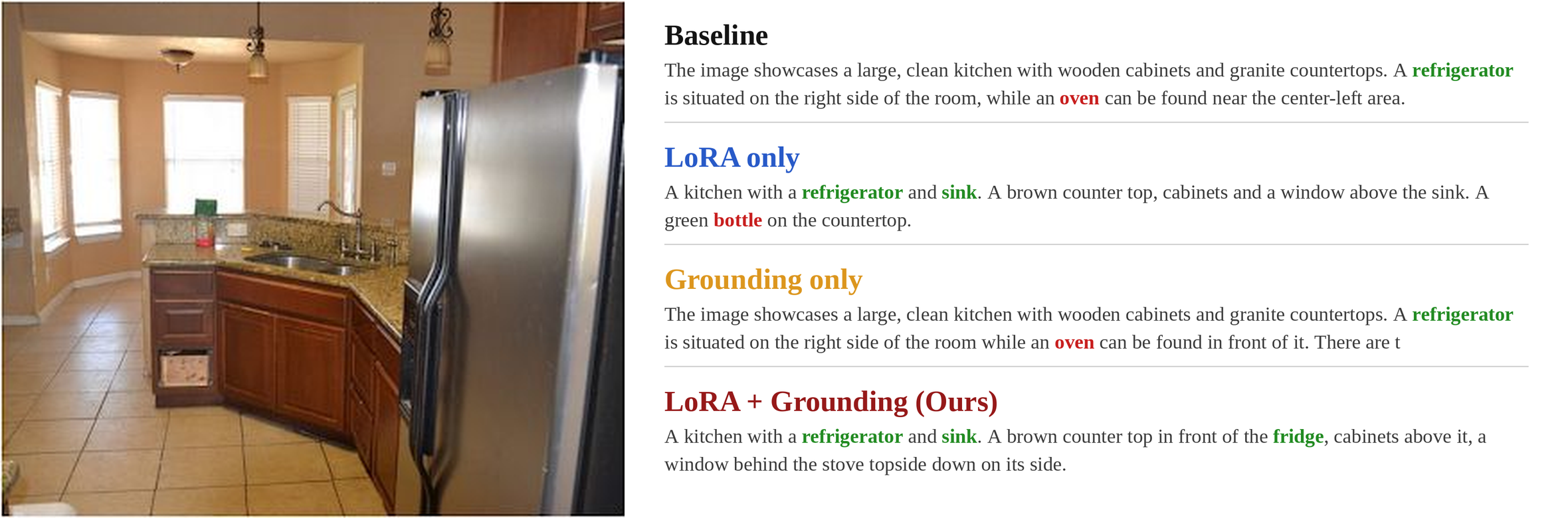}
  \caption{
  Qualitative comparison on a held-out kitchen image. Green object mentions are
  supported by COCO annotations; red mentions are counted as hallucinated by
  CHAIR. The combined LoRA + grounding method suppresses the oven hallucination
  while retaining grounded objects, but yields a shorter, more conservative
  caption.
  }
  \label{fig:qualitative-example}
\end{figure}

Figure~\ref{fig:qualitative-example} shows a held-out kitchen example. The
baseline correctly mentions the refrigerator but hallucinates an unsupported
oven, and the grounding-only method repeats the same hallucination. LoRA-only
removes the oven but introduces another unsupported object mention. The combined
LoRA + grounding method retains grounded objects such as the refrigerator and
sink while suppressing the oven hallucination, but produces a shorter and less
fluent caption. This mirrors the aggregate pattern: fewer unsupported object
mentions, with a tradeoff toward more conservative captions.

\section{Discussion}
Object-level diagnosis can be converted into an actionable intervention. The
ablation-screened 32-head set supports both a targeted adapter and an
inference-time control signal, and the full method gives the strongest paired
reduction in hallucination. The random-head control
(Section~\ref{sec:random-head}) suggests that this leverage depends on the
selected heads rather than LoRA capacity alone, while the LoRA-scale sweep shows
that the grounding controller adds value across adapter strengths.

The main tradeoff is that hallucination reduction comes with more conservative
generation. CHAIR measures an important object-level failure mode, but reducing
unsupported mentions can also reduce informativeness and object coverage. The
fixed-budget analysis (Section~\ref{sec:budget-robustness}) addresses the
strongest version of the length-artifact concern: the CHAIR reduction persists
when methods share the same token budgets and grows as more tokens are allowed.
At the same time, the method still produces shorter captions within those
budgets. We therefore interpret the result as targeted hallucination reduction,
not as a uniform improvement in caption quality.

\section{Limitations and Future Work}
This study focuses on one model family, one captioning prompt, and COCO-style
object annotations. CHAIR provides a useful controlled signal, but it depends on
annotation coverage and synonym matching, so valid but unannotated objects may be
counted as hallucinations. The grounding score also uses attention as a
lightweight proxy for visual grounding, even though attention weights are not
guaranteed to be faithful explanations of model behavior~\citep{jain2019attention,
wiegreffe2019attention}, and the selected heads are screened by ablation.

Several follow-up controls would sharpen the attribution. A fully length-matched
comparison could force the baseline to match the average output length of our
method, separating reduced hallucination from reduced object coverage more
directly. An all-head LoRA baseline and a plain DPO baseline at matched length
would test whether head restriction gives advantages beyond targeted capacity
control. Human evaluation would also help measure fluency, informativeness, and
faithfulness beyond CHAIR. Finally, extending the pipeline to additional VLMs,
prompts, and datasets would test whether hallucination-linked attention heads
provide stable intervention targets beyond LLaVA on COCO captions.

\section{Conclusion}
We presented a compact diagnosis-to-intervention pipeline for object
hallucination in LLaVA captions: identify hallucination-linked attention heads,
adapt only those head slices, and add a grounding-aware decoding controller over
the same set. The full method significantly reduces CHAIRs and CHAIRi on 400
held-out COCO images (both $p < 0.001$), while a layer-matched random-head
control supports the relevance of the selected heads rather than LoRA capacity
alone. The budget analysis further shows that the reduction persists under fixed
decoding budgets and grows as more tokens are generated.

The resulting captions are shorter and more conservative, so the gains are best
understood as targeted reductions in object hallucination rather than as a
general improvement in caption quality. Still, the result is encouraging for
mechanistic interpretability in vision-language models: an internal diagnostic
signal can identify intervention sites with measurable behavioral leverage. More
broadly, object hallucination offers a useful testbed for moving from
interpretability as post-hoc explanation toward interpretability as a guide for
localized model editing and controlled behavioral change.

\bibliographystyle{abbrvnat}
\bibliography{references}

\end{document}